\documentclass[letterpaper]{article} 
\usepackage{aaai25}  
\usepackage{times}  
\usepackage{helvet}  
\usepackage{courier}  
\usepackage[hyphens]{url}  
\usepackage{graphicx} 
\usepackage{natbib}  
\usepackage{caption} 
\usepackage{algorithm}
\usepackage{algorithmic}

\usepackage{makecell} 

\usepackage{amssymb}
\usepackage{amsmath}
\usepackage{color}
\usepackage{cleveref}
\usepackage{multirow}
\usepackage{booktabs}
\usepackage{xspace}
\usepackage[table]{xcolor}
\usepackage{subcaption}
\definecolor{mygray}{gray}{.9}
\definecolor{ForestGreen}{RGB}{34,139,34}
\usepackage{amsthm}

\newcommand{\train}{\mathcal{D}}
\def\BR{{\mathbb{R}}}
\renewcommand{\vec}[1]{\boldsymbol{#1}}

\newcommand{\norm}[1]{\left\Vert #1 \right\Vert}

\makeatletter
\DeclareRobustCommand\onedot{\futurelet\@let@token\@onedot}
\def\@onedot{\ifx\@let@token.\else.\null\fi\xspace}

\def\eg{\emph{e.g}\onedot} 

\def\ie{\emph{i.e}\onedot}

\def\etc{\emph{etc}\onedot}

\def\etal{\emph{et al}\onedot}
\makeatother
\newcommand{\name}[0]{NCKD\xspace}

\usepackage{newfloat}
\usepackage{listings}
\DeclareCaptionStyle{ruled}{labelfont=normalfont,labelsep=colon,strut=off} 
\floatstyle{ruled}
\newfloat{listing}{tb}{lst}{}
\floatname{listing}{Listing}

\usepackage{xcolor}

\title{Neural Collapse Inspired Knowledge Distillation}

\author {
    Shuoxi Zhang,
    Zijian Song,
    Kun He\thanks{Corresponding Author}
}
\affiliations {
    School of Computer Science and Technology, Huazhong University of Science and Technology\\
    \{zhangshuoxi,songzijian88,brooklet60\}@hust.edu.cn,
}

\usepackage{bibentry}

\begin{document}

\maketitle

%

\begin{abstract}
Existing knowledge distillation (KD) methods have demonstrated their ability in achieving student network performance on par with their teachers. However, the knowledge gap between the teacher and student remains significant and may hinder the effectiveness of the distillation process. In this work, we introduce
 the
structure of Neural Collapse (NC) into the KD framework. NC typically occurs in the
final phase of training, resulting in a graceful geometric structure where the last-layer features form a simplex equiangular tight frame. Such phenomenon has improved the generalization of deep network training. 
We hypothesize that NC can also alleviate the knowledge gap in distillation, thereby enhancing student performance. This paper begins with an empirical analysis to bridge the connection between knowledge distillation and neural collapse. Through this analysis, we establish that transferring the teacher's NC structure to the student benefits the distillation process. Therefore, instead of merely transferring instance-level logits or features, as done by existing distillation methods, we encourage students to learn the teacher's NC structure. Thereby, we propose a 
new distillation paradigm termed Neural Collapse-inspired Knowledge Distillation (NCKD). Comprehensive experiments demonstrate that NCKD is simple yet effective, 
improving the generalization of all distilled student models and achieving state-of-the-art accuracy performance.

\end{abstract}

\begin{figure}[t]
\centering
\includegraphics[width=0.54\columnwidth]{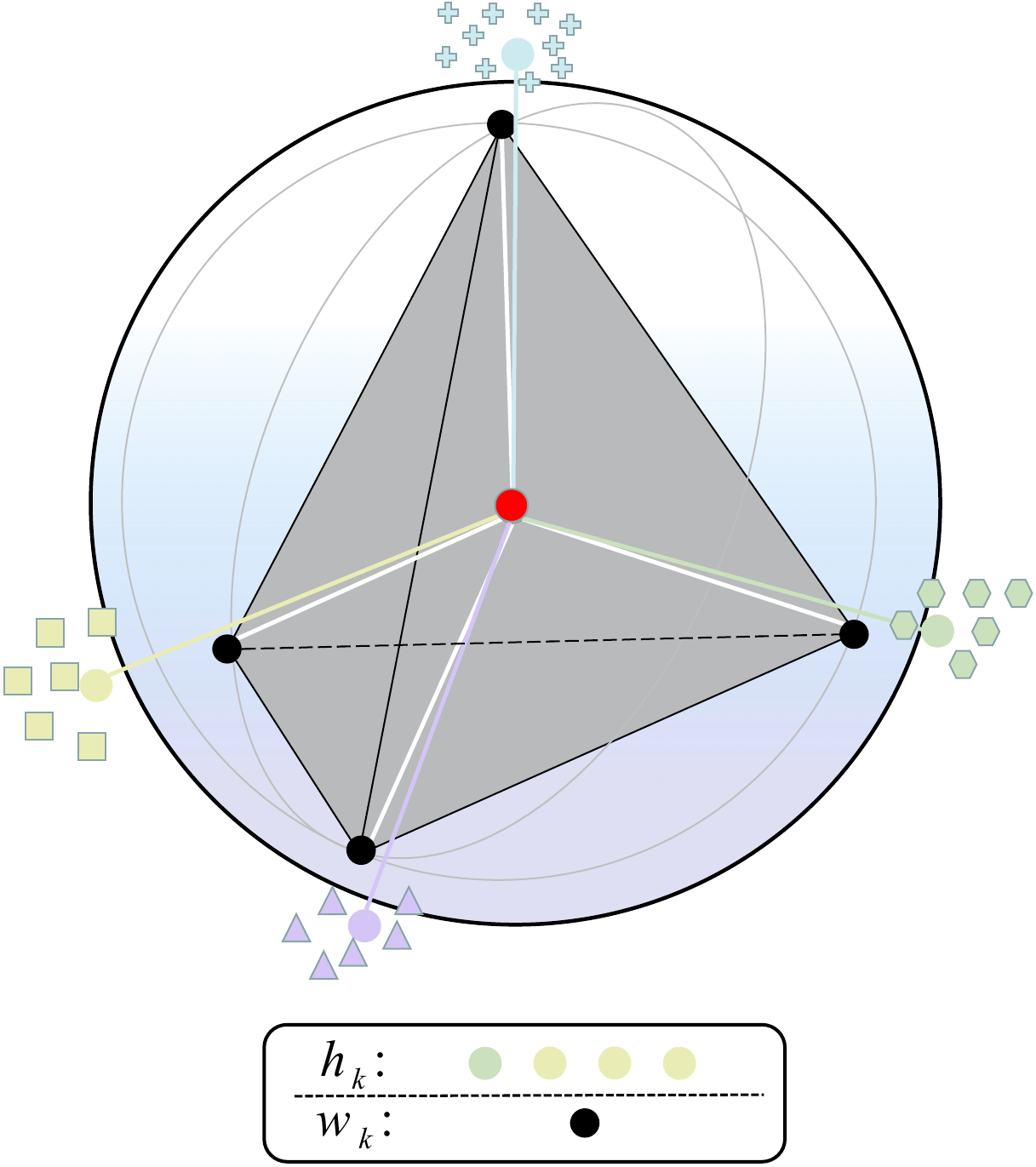} 
\caption{\small Description of the structure of Neural Collapse. All class features progressively collapse toward their centroids, forming an equiangular, elegant structure. Also, classifier $\Vec{w}$ will align with its corresponding last-layer normalized centroid $\Tilde{\Vec{h}}$.}
\label{fig:fig_1}

\end{figure}

\section{Introduction}
{I}{n} recent decades, deep learning has made remarkable strides in the field of computer vision, resulting in significant advancements in performance and generalization across various downstream tasks, including image classification~\cite{resnet,resnetv1,hu2018squeeze, ViT}, object recognition~\cite{girshick2015fast,lin2017feature,chen2019mmdetection}, and semantic segmentation~\cite{zhao2017pyramid, poudel2019fast}, \etc

These remarkable achievements  
have been largely attributed to the effectiveness of over-parameterized networks. However, the cumbersome deep models typically require substantial computation and memory resources during training and inference stages, making it challenging to deploy on mobile devices or embedded systems with limited resources. To address this issue, knowledge distillation~\cite{KD} (KD) has emerged as a crucial technique for model compression and performance improvement. By transferring knowledge encapsulated in a large, well-trained teacher model to a smaller student model, KD aims to achieve comparable performance in a more resource-efficient manner. This process is particularly beneficial in scenarios where deploying large models is impractical due to computational constraints. Despite its widespread adoption, the efficacy of KD is often limited by a persistent knowledge gap between the teacher and student models, resulting in suboptimal student performance.



Meanwhile, a parallel line of research has uncovered the phenomenon of \texttt{Neural Collapse}~\cite{papyan2020NC} (NC), where the final layer representations of a deep neural network exhibit a surprisingly symmetric and structured geometry as training progresses. Neural collapse is characterized by the alignment of within-class feature vectors, which converge to their respective class means, forming a simplex equiangular tight frame (ETF) (see \Cref{fig:fig_1}). The occurrence and prevalence of NC have been empirically verified through experiments with various datasets and network architectures\cite{NCinMSE}. This phenomenon not only contributes to model interpretability but also enhances its generalization capabilities.

The study of NC arguably provides a better understanding on the properties of deep features. Nonetheless, existing research has not addressed the following questions: 
\emph{ Are desirable KD methods the result of inducing a better simplex ETF structure? Can we improve the distillation process by encouraging the student to learn the teacher's NC structure?} 

Given the geometric elegance and generalization benefits of neural collapse, we hypothesize that integrating these properties into the student model can bridge the knowledge gap more effectively. 
Thus, we strive to investigate whether existing KD techniques enable the student model to obtain the NC structure of the teacher and leverage this phenomenon to enhance KD performance.

In this paper, we first conduct an empirical analysis to explore the relationship between the student's NC structure and its impact on the distillation process. Through this analysis, we establish that a well-aligned NC structure, which enhances generalization in plain training, also plays a crucial role in bridging the knowledge gap and improving performance within the KD paradigm. Accordingly, we  exploit the properties of $\mathcal{NC}_\mathbf{1}$, where the features of a well-trained network collapse towards their respective class centroids. 
We design a contrastive loss that encourages the student's feature space to align closely with the teacher's centroids. Next, we extend this approach by transferring the teacher’s ETF structure to the student, ensuring that the student’s class not only aligns with the corresponding teacher centroids but also forms a consistent ETF structure relative to other classes. Finally, considering that the primary goal of KD is to reduce computational costs, we capitalize on the properties of $\mathcal{NC}_\mathbf{3}$ by using normalized prototypes as the classifier, thereby reducing computational overhead. The above three key components form the foundation of our {\bf N}eural  {\bf C}ollapse-inspired  {\bf K}nowledge  {\bf D}istillation (NCKD) framework.

We conduct extensive experiments to evaluate the effectiveness of NCKD across various benchmarks. Our method not only outperforms state-of-the-art distillation techniques on multiple vision tasks but also demonstrates its versatility as a plug-and-play loss that can be integrated into other popular distillation methods to enhance their performance. 

Our main contributions can be summarized as follows:
\begin{itemize}

    \item We explore the intersection of two intensively studied fields, knowledge distillation and neural collapse, and attempt to establish a connection. To the best of our knowledge, we are the first to apply the principles of NC within the KD framework.
\item We distill the teacher's NC structure into the student model. Our approach goes beyond merely distilling class semantics; more critically, we also distill the ETF structure formed by the classes, thereby encouraging the student to construct a similarly elegant structure as that of the teacher.
    \item Our approach consistently outperforms state-of-the-art baselines in extensive experiments, encompassing various network architectures and diverse tasks including classification and detection.
\end{itemize}

\begin{figure*} [t]
\vspace{-25pt}
\centering
\subfloat[$\mathcal{NC}_\mathbf{1}$]{
    \includegraphics[scale=0.27]{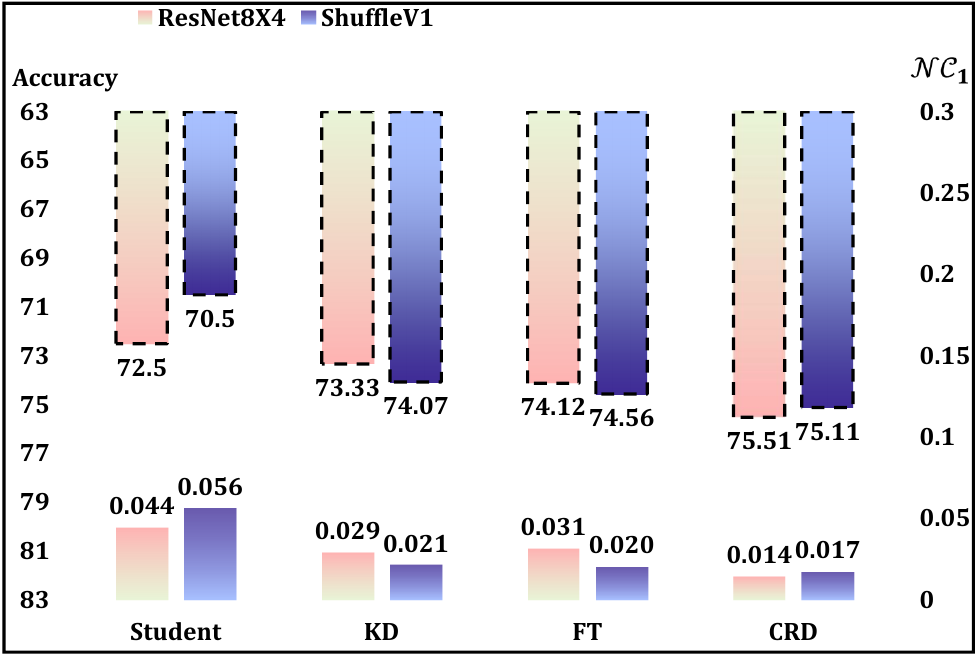}
    
    }
\subfloat[$\mathcal{NC}_\mathbf{2}$]{
    \includegraphics[scale=0.27]{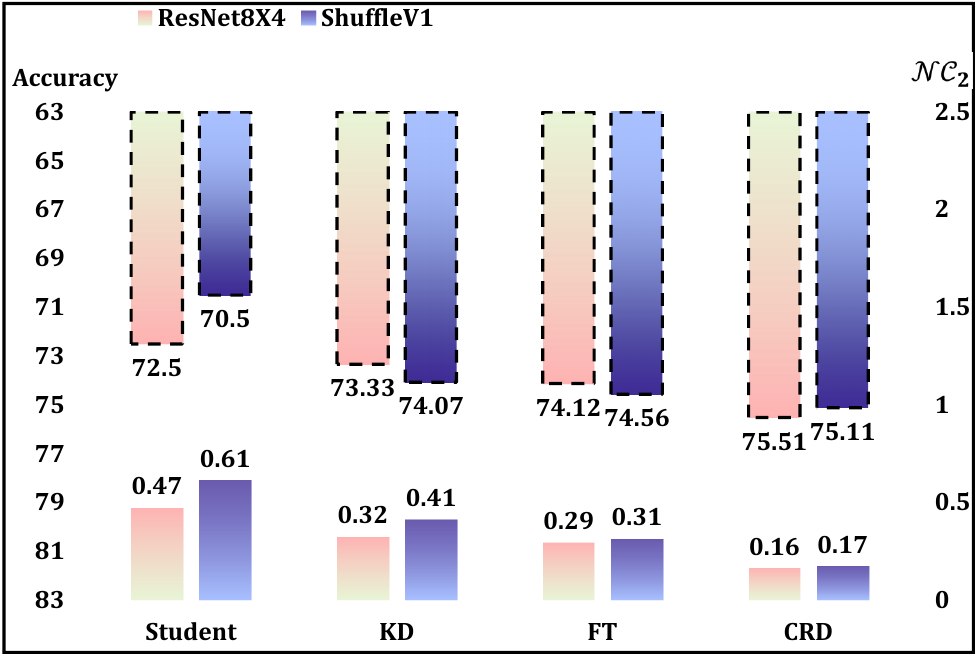}

    }
\subfloat[$\mathcal{NC}_\mathbf{3}$]{
    \includegraphics[scale=0.27]{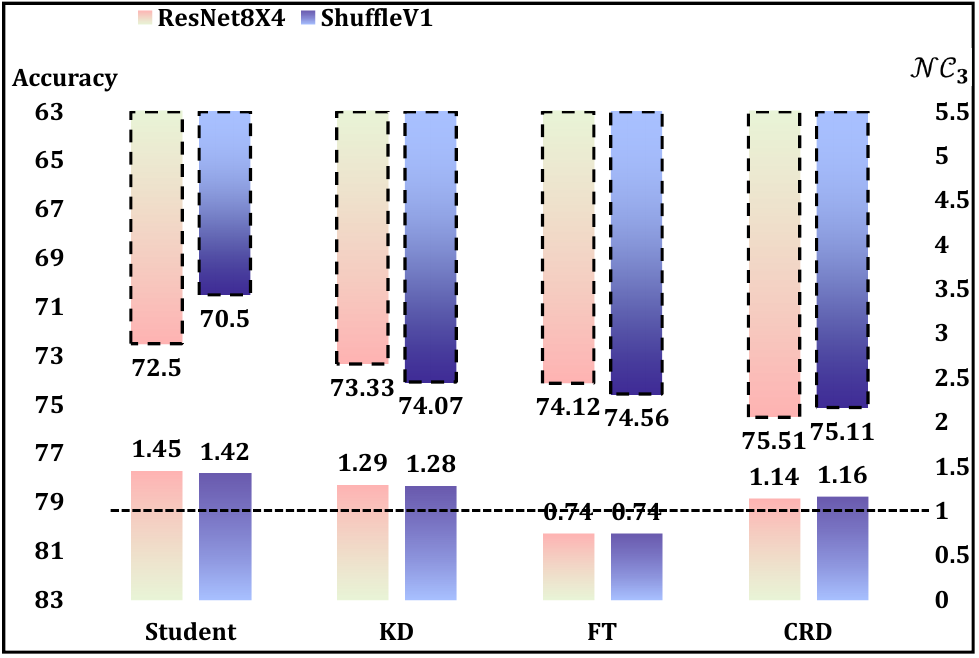}
    }
    \vspace{-5pt}
\caption{\small Comparison of NC metrics and prediction performance across different methods. Both networks were distilled from ResNet32x4 on CIFAR-100. The ideal NC results are characterized by $\mathcal{NC}_\mathbf{1,2}$ approaching 0, and $\mathcal{NC}_\mathbf{3}$ approaching 1.}
\label{fig: accuracy_NC} 
\vspace{-10pt}
\end{figure*}

\section{Related Work}
In this section, we first provide a brief overview of the related studies on knowledge distillation, including several state-of-the-art methods. Following that, we review the research literature on neural collapse and discuss its applications in various specific domains.

\subsection{Knowledge Distillation}
Knowledge distillation~\cite{KD} was first introduced by Hinton \etal, who utilized dark knowledge hidden within the well-trained teacher network to improve the performance of the student. They employed the probabilistic relationships from the negative logits to provide additional supervision and better regularization~\cite{CS-KD}. Building on this, logit-based distillation has demonstrated its potential in improving student model performance and generalization. Subsequent works have further refined logit-based KD through structural information~\cite{RKD, CCKD} or graph-level knowledge~\cite{IRG, TRG}. However, a significant knowledge gap persists between teacher and student models, prompting researchers to explore more effective knowledge transfer methods. For example, Kim \etal~\cite{kim2021comparing} proposed relaxing the KL divergence constraint~\cite{KLDivergence} to enhance information transfer, while Zhao \etal~\cite{DKD} decoupled traditional KD loss to achieve more efficient and adaptable distillation.

Another line of KD research leverages information concealed in intermediate features, attempting to align the feature maps between the teacher and student. FitNet~\cite{Fitnets} initiated this line by mimicking the teacher's intermediate features, setting the stage for feature-based distillation. Subsequent methods have refined the alignment and knowledge transfer from teacher features, incorporating attention mechanisms~\cite{AT, CAT-KD}, neural selectivity~\cite{NST}, and specifically designed alignment modules~\cite{FT, CKD, ReviewKD, TTM}.

\subsection{Neural Collapse}
\texttt{Neural collapse} (NC) refers to a phenomenon where the features and classifiers of a neural network’s final layer progressively converge to form a simplex \emph{equiangular tight frame} (ETF), an elegant geometric structure. Empirical evidence of NC has been observed with both cross-entropy loss~\cite{papyan2020NC,NCinCE,zhu2021UFM} and mean squared error (MSE) loss~\cite{NCinMSE,mixon2022neural}. This phenomenon is pervasive in deep training, arising unbiased to disparate datasets or architectures. Consequently, it is observed in nearly all standard classification tasks, including those involving imbalanced datasets~\cite{dang2023imbalanced}. Conceptually, NC represents the network's goal to maximize inter-class distances, thereby enhancing both generalization and adversarial robustness~\cite{papyan2020NC}. Consequently, NC has been effectively employed to improve performance in areas such as contrastive learning~\cite{xue2023features}, class incremental learning~\cite{yang2023neural,seo2024learning,kimfixed}, and out-of-distribution (OOD) detection~\cite{ammar2023neco}. However, the manifestation of NC in knowledge distillation, and its potential integration into distillation strategies, remain largely unexplored.


\begin{figure*}[t]
\vspace{-35pt}
\centering
\includegraphics[width=0.73\linewidth]{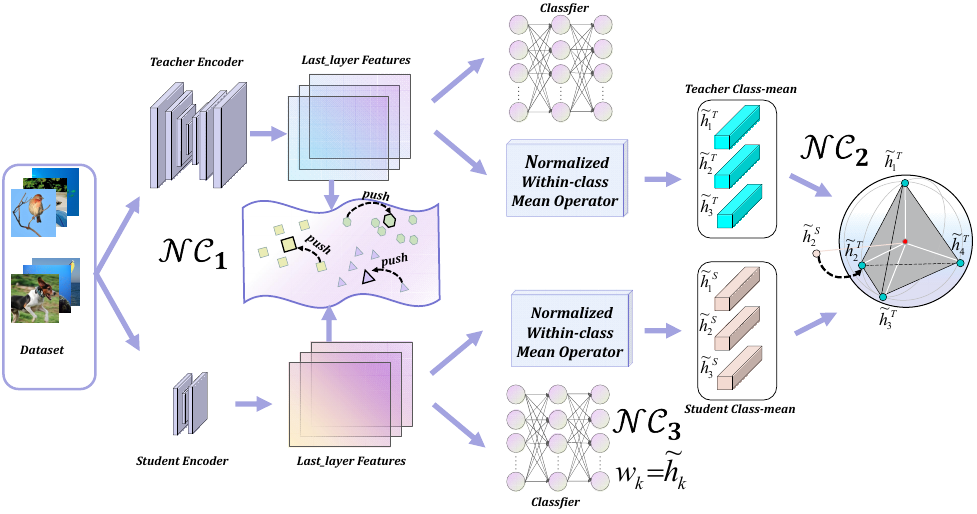} 
\vspace{-5pt}
\caption{\small The overall framework of our \name. We distill the $\mathcal{NC}_\mathbf{1,2}$ from the teacher to the student. We normalize within-class mean $\vec{h}$ to $\Tilde{\vec{h}}$ to construct the ETF structure.  illustrate $\mathcal{NC}_\mathbf{2}$ distillation using $\Tilde{\vec{h}}^S_2$ as the example, which replicates the teacher's ETF structure with other classes. $\mathcal{NC}_\mathbf{3}$ classifier is leveraged to reduce computational costs.}
\label{fig:fig_Framework}
\vspace{-10pt}
\end{figure*}

\section{Problem Formulation}
In this section, we first introduce several fundamental KD methods for subsequent analysis and provide  necessary notations to facilitate the ensuing explanations. We then provide an overview of neural collapse, outlining its core properties and the metrics used to characterize this phenomenon. Finally, we empirically examine the impact of neural collapse on the generalization of networks trained with various representative KD methods.

\subsection{Knowledge Distillation}
Consider the $K$-class classification problem
$\train=\{(\boldsymbol{x}_{k}^{(n)}, \boldsymbol{y}_k)\}_{ k \in[K],n\in[N_k]}$. Here $N_k$ is the number of samples in the $k$-th class. For simplicity, our distillation framework assumes a balanced dataset, meaning $N_k \equiv N$, resulting in a total dataset size of $N*K$. Each sample consists of a data point $\boldsymbol{x}_{k}^{(n)}$ and the one-hot label $\boldsymbol{y}_k\in \BR^K$. In addition, we utilize $\vec{f}_\ell$ and $\vec{z}$ to denote the intermediate feature from the $\ell$-th ($\ell \in [1,L]$) layer and the corresponding output logits, respectively. Specifically, we use ${g}$ to represent the feature function in the penultimate layer.
 
In the basic KD paradigm, knowledge from the teacher is encapsulated and transferred through prediction logits or intermediate features. The total distillation loss can be formulated as:
\begin{equation}\label{eq:kd loss}
\mathcal{L} = \mathcal{L}_\text{cls} + \alpha\mathcal{L}_\text{distill},
\end{equation}
where $\mathcal{L}_\text{cls}$ is the classification loss with ground-truth labels, and the term $\mathcal{L}_\text{distill}$ indicates the distillation loss. 

In Hinton's vanilla KD, it uses $\mathcal{L}_{KD}$ as $\mathcal{L}_\text{distill}$ to measure the KL divergence~\cite{KLDivergence} of softened logit predictions ($\boldsymbol{z}_S$, $\boldsymbol{z}_T$) between the teacher and the student:
\begin{equation}\label{eq:kl loss}
\mathcal{L}_{KD}={\tau}^2{KL}\big( \sigma(\boldsymbol{z}_S/\tau), \sigma(\boldsymbol{z}_T/\tau) \big), 
\end{equation}
where $\sigma$ denotes the Softmax operation, and the temperature $\tau$ is used to soften the logits. As $\tau$ increases, the probability becomes softer, enabling a more comprehensive encoding of the categorical relationships imparted by the teacher.

Beyond distilling knowledge from logits, valuable information is also contained in intermediate features. Feature-based methods (\eg, FT~\cite{FT}) leverage the intermediate features from the teacher to guide the student's training. Accordingly, the distillation loss $\mathcal{L}_{FT}$ for $\mathcal{L}_\text{distill}$ is given by:
\begin{equation}
    \mathcal{L}_{FT} = \mathcal{D}\left(\Phi(f_L^T), F_L^S\right),
    \label{eq:feature loss}
\end{equation}
where $F^T_L$ and $F^S_L$ are the $L$-th (the last-layer) intermediate features of teacher and  student, respectively. $\mathcal{D}(\cdot)$ denotes the distance function, utilized to measure the discrepancy of the selected features and thereby guide the distillation process. Extra transformation layer $\Phi$ is used to align the feature sizes between teacher and student.


\subsection{Neural Collapse} \label{sec:NC formulation}
Neural collapse constructs an elegant geometric structure on the last-layer feature and the classifier in the {final} training phase. For simplicity, we denote the last-layer feature $g(\vec{x}_k^{(n)})$ of the sample $\vec{x}_k^{(n)}$ by $\vec{h}_k^{(n)}$. And the $k$-th \emph{class means} and \emph{global mean} of the features are calculated by:$$
\vec{h}_k := \frac{1}{N}\sum_{i=1}^{N} \vec{h}_k^{(n)}, \qquad
\vec{h}_G := \frac{1}{K}\sum_{k=1}^K \vec{h}_k.
$$The NC phenomenon includes the following properties:



\begin{enumerate}
    \item
    \textbf{NC1: Within-class variability collapse.} $\mathcal{NC}_\mathbf{1}$ depicts the relative magnitude of within-class variability $\mathbf{\Sigma_W} = \frac{1}{NK} \sum_{k=1}^K\sum_{n=1}^N(\vec{h}_k^{(n)}-\vec{h}_k)(\vec{h}_k^{(n)}-\vec{h}_k)^\top$ in relation to the total variability. We compute $\mathcal{NC}_\mathbf{1}$ by using within-class covariance $\mathbf{\Sigma_W}$ and between-class covariance $\mathbf{\Sigma_B} = \frac{1}{K}\sum_{k=1}^K (\vec{h}_k - \vec{h}_G)(\vec{h}_k - \vec{h}_G)^\top$. Thus, we can measure the $\mathcal{NC}_\mathbf{1}$ collapse by measuring the magnitude of the between-class covariance $\mathbf{\Sigma_B} \in \BR^{d\times{d}}$  compared to the within-class covariance $\mathbf{\Sigma_W} \in \BR^{d\times{d}}$ of the learned features via:
\begin{small}
    \begin{align}\label{eq:NC1}
    \mathcal{NC}_\mathbf{1}\;:=\;\frac{1}{K}\mathrm{Trace}\left(\mathbf{\Sigma_W}\mathbf{\Sigma_B}^\dagger\right),
    \end{align}
\end{small}
where $\mathbf{\Sigma_B}^\dagger$ denotes the pseudo inverse of $\mathbf{\Sigma_B}$.
    
    \item
    \textbf{NC2: Convergence to Simplex ETF.} The penultimate feature centroids exhibit a simplex ETF structure with the following property: if we define the normalized class means as $\tilde{\vec{h}}_k = \frac{\vec{h}_k - \vec{h}_G}{\norm{\vec{h}_k - \vec{h}_G}_2}$, then $\langle{\tilde{\vec{h}}_k,\tilde{\vec{h}}_{k'}\rangle} = -\frac{1}{K-1}$ for $k \neq k'$,    
    indicating that the centered class means are equiangular.  Then we define the $\mathcal{NC}_\mathbf{2}$ as:
\begin{small}
    \begin{equation}
\mathcal{NC}_\mathbf{2}=\operatorname{avg}_{k\neq k'}\left(\left| {\langle{\tilde{\vec{h}}_k}, \tilde{\vec{h}}_{k'}\rangle} + \frac{1}{K-1}\right|\right).        
    \end{equation} 

\end{small}

    \item
    \textbf{NC3: Convergence to self-duality.}
    The within-class means centered by the global mean will be {aligned with} their corresponding classifier weights, which means the classifier weights will {converge to} the same simplex ETF:
    \begin{small}
        \begin{equation}
        \mathcal{NC}_\mathbf{3} = \operatorname{avg} \norm{\frac{\langle{\tilde{\vec{h}}_k, \vec{w}_k}\rangle}{\norm{\tilde{\vec{h}}_k}\cdot\norm{\vec{w}_k}}}_F.
        \label{eq:nc3}
    \end{equation}
    \end{small}

\end{enumerate}

We evaluate the student model’s last-layer feature and classifier under different training conditions --- namely, standalone student training, KD, FT, and CRD~\cite{CRD} --- and compare the resulting NC metrics with their respective distillation performance (as shown in Figure \ref{fig: accuracy_NC}). In both distillation pairs, a strong correlation between NC and distillation outcomes is evident. Improved distillation often corresponds with decreases in $\mathcal{NC}_\mathbf{1}$ and $\mathcal{NC}_\mathbf{2}$. Among the methods, CRD achieves the best distillation results, with $\mathcal{NC}_\mathbf{1}$ and $\mathcal{NC}_\mathbf{2}$ values closest to zero and $\mathcal{NC}_\mathbf{3}$ closest to one. This indicates that the distillation process may implicitly steer the student toward an optimal NC structure. Thus, directly leveraging NC properties in distillation would be a highly effective strategy.

\section{The Proposed Method}
Building on the relationship between NC and KD, we propose an NC-inspired distillation method that explicitly promotes NC-like behavior in the student model. Our approach comprises three key components: 1) a contrastive learning module that aligns the student with the teacher's prototypes; 2) a mechanism to distill the teacher's neural ETF structure into the student; and 3) a $\mathcal{NC}_\mathbf{3}$ classifier designed to reduce computation.  The overall framework is shown in Figure \ref{fig:fig_Framework}.

\subsection{$\mathcal{NC}_\mathbf{1}$ Distillation}
In the above analysis, we have established the 
${\mathcal{NC}_\mathbf{1}}$ 
  property of a well-trained network, indicating that the last-layer features exhibit reduced within-class variance, effectively collapsing to their respective class centroids. This naturally leads to the idea of directly aligning the student features with the teacher's corresponding prototypes. To achieve this alignment, we leverage the paradigm of contrastive learning, which has already demonstrated its ability to preserve the NC phenomenon~\cite{kini2023supervised}. We introduce the prototype alignment loss as follows:

\begin{small}
\begin{equation}
    \mathcal{L}_{{\mathcal{NC}_\mathbf{1}}} = -\frac{1}{NK} \sum_{n,k}^{NK} \log \frac{\exp\left(\text{sim}\left(g^S(x_k^{(n)}), \Vec{h}_k^{T}\right)/\tau\right)}{\sum_{k=1}^{K} \exp\left(\text{sim}\left(g^S(x_k^{(n)}, \Vec{h}_k^{T}\right)/\tau\right)}.
\end{equation}    
\end{small}
Here, $\tau$ is the temperature parameter that controls the feature space structure, and $\text{sim}(\cdot)$ denotes the similarity measure. To address the norm gap between teacher and student, as discussed in \cite{wang2023ND}, we use standard cosine similarity, $\cos(\Vec{a},\Vec{b}) = \frac{\Vec{a}\cdot\Vec{b}}{|\Vec{a}|\cdot|\Vec{b}|}$, to quantify the disparity between student features and their corresponding teacher centers.

The CRD loss~\cite{CRD} is most closely related to our approach, as it also employs a contrastive framework to enhance distillation matching. However, the key difference lies in the alignment strategy: while CRD aligns teacher and student features on an instance-wise basis, our method directly aligns student features with the teacher's prototypes. This design choice is driven by the observation that a well-trained teacher’s features naturally collapse toward class centers, reflecting the ${\mathcal{NC}_\mathbf{1}}$ property. In the experimental section, we will compare the effects of the two loss functions.


\subsection{$\mathcal{NC}_\mathbf{2}$ Distillation}
To fully leverage the structured feature space of a well-trained teacher model, it is essential to distill the simplex ETF structure into the student model.  As described earlier, the modified within-class feature means $\Tilde{\Vec{h}}_k$ collectively form an equiangular fabric. For simplicity, we organize all prototypes of the student and teacher into matrices $\Tilde{\Vec{H}}^S, \Tilde{\Vec{H}}^T  \in \BR^{K\times{D}}$,  where each row represents the corresponding class mean. We aim to ensure that each student's normalized centroid
 $\Tilde{\Vec{h}}_k^S$ mimics the ETF structure of the teacher, thereby preserving the inter-class relationships. To achieve this, we propose the following loss function:
\begin{small}
    \begin{equation}
        \mathcal{L}_{\mathcal{NC}_2} = \norm{\Tilde{\Vec{H}}^S ({\Tilde{\Vec{H}}^T})^\top - \frac{K}{K-1}\left(\Vec{I}_K - \frac{1}{K}{\bf 1}_K {{\bf 1}_K}^\top\right)}_2^2.
        \label{eq:NC2}
    \end{equation}
\end{small}
Here,  \( I_K \) represents the identity matrix of dimension \( K \), and \( \mathbf{1}_K \) denotes a vector of ones with \( K \) elements. Notably, the product \( \mathbf{1}_K \mathbf{1}_K^\top \) yields a \( K \times K \) matrix where all elements are equal to 1. Ideally, when the $\mathcal{L}_{\mathcal{NC}_\mathbf{2}}$ loss is optimized to $0$, each normalized centroid $\Tilde{\Vec{h}}_k^S$ of the student model will have a similarity score of $1$ with the corresponding teacher's centroid $\Tilde{\Vec{h}}_k^T$, while displaying an inner product of \(-\frac{1}{K-1}\) with the centroids of other classes, thus elegantly matching the teacher's simplex ETF structure\footnote{A detailed explanation of why \(\mathcal{L}_{\mathcal{NC}_\mathbf{2}}\) enforces a simplex ETF is provided in the Appendix.}. Consequently, optimizing this loss allows us to effectively distill the \(\mathcal{NC}_{\mathbf{2}}\) structural knowledge from teacher to student, ensuring that the student model accurately mimics the geometric configuration of the teacher's class centroids. The total loss in our framework can be formulated as:
    \begin{equation}
        \mathcal{L}_\text{total} = \mathcal{L}_\text{cls} + \lambda_1 \mathcal{L}_{\mathcal{NC}_\mathbf{1}} + \lambda_2 \mathcal{L}_{\mathcal{NC}_\mathbf{2}}, 
        \label{eq:total}
    \end{equation}
where $\lambda_1, \lambda_2$ are the balancing coefficients. 

\subsection{$\mathcal{NC}_\mathbf{3}$-inspired Classifier}
The primary goal of KD is to minimize computational costs in practical applications while maintaining the model performance. Given the previously discussed $\mathcal{NC}_\mathbf{3}$ property, where the final-layer features tend to form a self-dual space towards the end of the training phase --- meaning that the features of each class align closely with their corresponding functional (i.e., the classifier). Therefore, a natural idea is to eliminate the classifier computation. Instead, we utilize the normalized centroid to represent the corresponding classifier weight  \( \Vec{w} \) in the following form: 
\begin{small}
    \begin{equation*}
        \Vec{w}_k = \Tilde{\Vec{h}}_k
    \end{equation*}
\end{small}
This approach leverages the 
$\mathcal{NC}_\mathbf{3}$
  property to reduce computational overhead by eliminating the need for a separate linear classification layer. Notably, several existing distillation methods, such as SimKD~\cite{SimKD}, also implicitly utilize the 
$\mathcal{NC}_\mathbf{3}$
  property, though this is not always explicitly recognized in their design. We will explore this aspect further in our experimental section through a case study, providing additional insights.

  \begin{table*}
\vspace{-20pt}
    \centering
    \renewcommand{\arraystretch}{1.2}
        \resizebox{0.74\textwidth}{!}{%
        \begin{tabular}{c|lll|lll|c}
        \hline
        \multirow {3}{*}{Method} & \multicolumn{3}{c|}{Homogeneous architecture} & \multicolumn{3}{c|}{Heterogeneous architecture}& \multirow{3}{*}{Average} \\ 
        \cline{2-7}
        ~ & ResNet-56 & WRN-40-2 & ResNet-32$\times$4 & ResNet-50 & ResNet-32$\times$4 & ResNet-32$\times$4  \\
        ~ & ResNet-20 & WRN-40-1 & ResNet-8$\times$4 & MobileNet-V2 & ShuffleNet-V1 & ShuffleNet-V2 \\
        \hline
        {\tt teacher} (T) & 72.34 & 75.61 & 79.42 & 79.34 & 79.42 & 79.42 & 77.59\\
        {\tt student} (S) & 69.06 & 71.98 & 72.50 & 64.60 & 70.50 & 71.82 & 70.08\\
        \hline
        \multicolumn{8}{c}{\textit{Logit-based Method}}\\
        KD    & 70.66 & 73.54 & 73.33 & 67.65 & 74.07 & 74.45 & 72.28\\
        DKD      & 71.97 & 74.81 & 75.44 & 70.35 & 76.45 & 77.07 & 74.34\\
        DIST     & 71.78 & 74.42 & 75.79 & 69.17 & 75.23 & 76.08 & 73.75\\
        MLKD     & 72.19 & 75.35 & 76.98 & 69.58 & 77.18 & {\bf 77.92} & 74.87\\
        \hline
        \multicolumn{8}{c}{\textit{Feature-based Method}}\\
        FitNet   & 69.21 & 72.24 & 73.50 & 63.16 & 73.59 & 73.54 & 70.87\\
        RKD      & 69.61 & 72.22 & 71.90 & 64.43 & 72.28 & 73.21 & 70.61\\
        PKT     & 70.34 & 73.45 & 73.64 & 66.52 & 74.10 & 74.69 & 72.12\\
        CRD	     & 71.16 & 74.14 & 75.51 & 69.11 & 75.11 & 75.65 & 73.45\\
        ReviewKD & 71.89 & 75.09 & 75.63 & 69.89 & 77.45 & 77.78 & 74.62\\
        NORM	     & 71.55 & 74.82 & 76.49 & 70.56 & 77.42 & {77.87} & 74.79\\
        SimKD	     & 71.68 & 75.56 & {77.22} & 70.32 & 77.11 & {75.42} & 74.55\\
        TopKD	     & 71.58 & 74.43 & 75.40 & 69.12 & 75.04 & 76.33 & 73.65\\
        TTM	     & 71.83 & 74.32 & 76.17 & 69.24 & 74.18 & 76.52 & 73.71\\
        \rowcolor{mygray} \name & {\bf \color{ForestGreen}72.63 } & { \color{ForestGreen}75.71 }  & {{\color{ForestGreen}77.23 }} & {\color{ForestGreen}70.12 } & {\color{ForestGreen}77.48 } & {\color{ForestGreen}77.42 } & {\color{ForestGreen}75.10 }\\
        
        \hline
         CRD+\name	     & 72.26($\uparrow$1.10) & 75.16($\uparrow$1.02) & 76.88($\uparrow$2.74) & 69.88($\uparrow$0.77) & 76.32($\uparrow$1.21) & 76.68($\uparrow$1.03)  & {\bf 75.53($\uparrow$2.08)}\\
           SimKD+\name	     & 72.47($\uparrow$0.79) & {\bf 75.81($\uparrow$0.25)} & {\bf 78.18($\uparrow$0.94)} & {\bf 70.67($\uparrow$0.35)} & {\bf 77.71($\uparrow$0.60)} & 76.98($\uparrow$1.56)  & 75.30($\uparrow$0.75)\\
        \hline
        \end{tabular}
    }
        \vspace{-1mm}
        \caption{\small  Benchmarking results (mean of three repeats) on the CIFAR-100. Methods are reported with top-1 accuracy (\%). $\uparrow$ indicates the improvement of our approach when incorporated into others. The best results are highlighted with $\textbf{bold}$. 
        }
        \label{tab:cifar100-sota}
        \vspace{-5pt}
\end{table*}

\begin{table*}[t]
\centering
\resizebox{0.84\linewidth}{!}{
\begin{tabular}{c|l|cc|cccccccc|c}
\toprule
 Student (Teacher) & Metric & Teacher & Student & FT & KD & SP   & CRD  & ReviewKD & DIST  & TTM & DisWOT & \name \\
\midrule
\multirow{2}{*}{ResNet18 (ResNet34)}&Top-1 & 73.31 & 69.75 & 70.70 & 70.66 & 70.62 & 71.17 & 71.61  & 71.88 &  72.09 &
72.08 &\textbf{72.44} \\
&Top-5 & 91.42 & 89.07 & 90.00 & 89.88 & 89.80  & 90.13 & 90.51 & 90.42 & 90.48 & 90.38 & \textbf{91.12}\\
\midrule
\multirow{2}{*}{MobileNet (ResNet50)}&Top-1 & 76.16 & 70.13 & 70.78 & 70.68 & 70.99  & 71.37 & 72.56 & 72.94& 73.09 & 73.22& \textbf{73.61} \\
&Top-5 & 92.86 & 89.49 & 90.50 & 90.30 & 90.61  & 90.41 & 91.00 & 91.12 & 90.77 & 90.22 &\textbf{91.56}\\
\bottomrule
\end{tabular}
}
\vspace{-1mm}
\caption{\small 
Evaluation results of baseline settings on ImageNet. We use ResNet34 and ResNet50 as our teacher network.
}
\label{Tbl:ImageNet}
\end{table*}
\begin{table*}[t]
\vspace{-30pt}
\centering
\resizebox{0.75\linewidth}{!}{
\begin{tabular}{cl|ccc|ccc|ccc}
\toprule
& &  mAP & AP$_{50}$ & AP$_{75}$ & mAP & AP$_{50}$ & AP$_{75}$ & mAP & AP$_{50}$ & AP$_{75}$ \\
\midrule
\multirow{4}{*}{Method} & \multirow{2}{*}{Teacher} & \multicolumn{3}{c|}{ResNet101} & \multicolumn{3}{c|}{ResNet101} & \multicolumn{3}{c}{ResNet50} \\
&  & 42.04 & 62.48 & 45.88 & 42.04 & 62.48 & 45.88 & 40.22 & 61.02 & 43.81 \\
 & \multirow{2}{*}{Student} & \multicolumn{3}{c|}{ResNet18} & \multicolumn{3}{c|}{ResNet50} & \multicolumn{3}{c}{MobileNetV2} \\
 &&  33.26 & 53.61 & 35.26 & 37.93 & 58.84 & 41.05 & 29.47 & 48.87 & 30.90 \\
 \midrule
\multirow{3}{*}{Feature} & FitNet & 34.13 & 54.16 & 36.71 & 38.76 & 59.62 & 41.80 & 30.20 & 49.80 & 31.69 \\
&FGFI & 35.44 & 55.51 & 38.17 & 39.44 & 60.27 & 43.04 & 31.16 & 50.68 & 32.92 \\
& ReviewKD & 36.75 & 56.72 & 34.00 & 40.36 & 60.97 & 44.08 & 33.71 & 53.15 & 36.13 \\
\midrule
\multirow{5}{*}{Logits} & KD & 33.97 & 54.66 & 36.62 & 38.35 & 59.41 & 41.71 & 30.13 & 50.28 & 31.35 \\
&DIST & 34.89 & 56.32 & 37.68 & 39.24 & 60.82 & 42.77 & 31.98 & 52.33 & 34.02 \\
&DKD & 35.05 & 56.60 & 37.54 & 39.25 & 60.90 & 42.73 & 32.34 & 53.77 & 34.01 \\
\rowcolor{mygray}& {\bf \name} (Ours) & {\bf \color{ForestGreen} 37.36} & {\bf \color{ForestGreen} 57.96} & {\bf \color{ForestGreen} 37.94} & {\bf \color{ForestGreen} 40.68} & {\bf \color{ForestGreen} 62.12} & {\bf \color{ForestGreen} 44.89} & {\bf \color{ForestGreen} 33.97} & {\bf \color{ForestGreen} 54.32} & {\bf \color{ForestGreen} 35.41} \\
\bottomrule
\end{tabular}}
\vspace{5pt}
\caption{{\small Comparison results on MS-COCO. We take Faster-RCNN~\cite{ren2015faster} with FPN~\cite{xie2017aggregated} as the backbones, and AP, AP$_{50}$, and AP$_{75}$ as the evaluation metrics. The original accuracy results of the teacher and student models are also reported.}}
\label{tab:mscoco_results}
\end{table*}
\begin{table}[h!]
\centering\resizebox{0.48\textwidth}{!}{
\begin{tabular}{lc|cc|ccc|c}
\toprule
Teacher & Student & Teacher & Student & KD & DIST & DKD & \textbf{\name} \\
\midrule
ResNet-34 & \multirow{4}{*}{ResNet-18} & 73.31 & \multirow{4}{*}{69.76} & 71.21 & 71.88 & 71.68 & \textbf{72.44} \\
ResNet-50 & ~ & 76.13 & ~ & 71.35 & 72.04 & 71.91& \textbf{72.56} \\
ResNet-101 & ~ & 77.37 & ~ & 71.09 & 72.01 & 72.05 & \textbf{72.71} \\
ResNet-152 & ~ & 78.31 & ~ & 71.12 & 72.06 & 72.03& \textbf{72.77} \\
\midrule
Swin-T & \multirow{3}{*}{ResNet-34} & 81.70 & \multirow{3}{*}{73.31} & 74.56 & 74.78 & 74.92 & \textbf{74.95} \\
Swin-S & ~ & 83.00 & ~ & 74.68 & 74.69 & 74.82& \textbf{75.01} \\
Swin-B & ~ & 83.48 & ~ & 74.59 & 74.75 & 74.84 & \textbf{75.05} \\
\toprule
\end{tabular}}
\caption{\small Performance of ResNet-18/34 on ImageNet distilled from different large teachers.}
\label{tab: different teachers}
\end{table}

\begin{table}[t]
	\centering
	\resizebox{1\linewidth}{!}{
		\begin{tabular}{l|cccccc}  
			\toprule
			\multirow{2}{*}{Module}& KD &  \multicolumn{3}{c}{Distillation}& \multirow{2}{*}{ResNet-8$\times$4} & \multirow{2}{*}{ShuffleV1}\\  
		&	& $\mathcal{NC}_\mathbf{1}$ & $\mathcal{NC}_\mathbf{2}$ & $\mathcal{NC}_\mathbf{3}$  & & \\
			\midrule
			Baseline &-&-&-&-&72.51& 70.50 \\
   
			KD &\checkmark&-&-&-&74.12&74.00 \\
   CRD &\checkmark&-&-&-&75.51&75.11 \\
   CRD+$\mathcal{NC}$ &-&-&\checkmark&\checkmark&76.88&76.32 \\
            w/o $\mathcal{NC}_\mathbf{2}$ &-&\checkmark &- &\checkmark&75.98&76.48\\
			w/o $\mathcal{NC}_\mathbf{3}$ &-&\checkmark&\checkmark &-&77.00&77.24\\
			Ours &-&\checkmark&\checkmark&\checkmark&77.23&77.48\\
   Ours+KD &\checkmark&\checkmark&\checkmark&\checkmark&\textbf{77.41}&\textbf{77.55}\\
   
			\bottomrule
	\end{tabular}}
    
	\caption{\small Ablation study on the $\mathcal{NC}$-inspired distillation components on CIFAR-100. The baseline denotes the student's plain training. In other cases, the knowledge from pre-trained ResNet-32$\times$4 is used for distillation. }
	\label{tbl:ablation}
\end{table}

\begin{table}[!htbp]
\centering
    \begin{subtable}[t]{0.20\textwidth}

\begin{small}
\begin{tabular}{c|cc|}
\multicolumn{1}{c|}{Method} & \multicolumn{1}{c}{top-1} & $\mathcal{NC}_\mathbf{1}$ \\ \Xhline{1.2pt}
         KD              & 70.66                     & 2.7e-2     \\
CRD                        & 71.17                     & 1.4e-2 \\
\name                        & {\bf 72.44}                     & 8.1e-3

\end{tabular}
\end{small}
\end{subtable}
\begin{subtable}[t]{0.17\textwidth}

\begin{small}
\begin{tabular}{cc|c}
 \multicolumn{1}{c}{top-1} & $\mathcal{NC}_\mathbf{3}$ & \multicolumn{1}{c}{Method}  \\ \Xhline{1.2pt}
                        70.66                     & 1.47 & KD     \\
 72.01                    & 1.11  & SimKD \\
 {\bf 72.44}                     & 1.07 & \name

\end{tabular}
\end{small}
\end{subtable}
\caption{\small We use ResNet34/ResNet18 pair training on ImageNet to test the implicit $\mathcal{NC}$ properties of some existing approaches. }
\label{tab:case study}
\end{table}
\begin{figure*} [t]
\vspace{-5pt}
	\centering
	\subfloat[KD]{
		\includegraphics[scale=0.21]{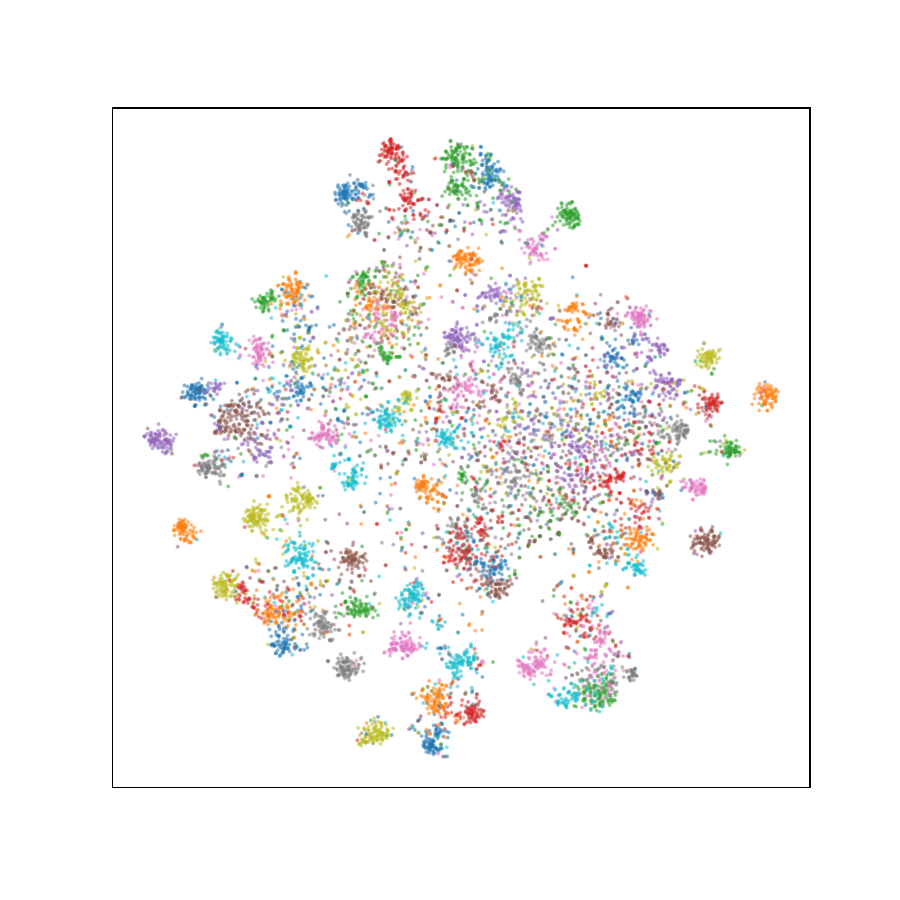}
		}
	\subfloat[CRD]{
		\includegraphics[scale=0.21]{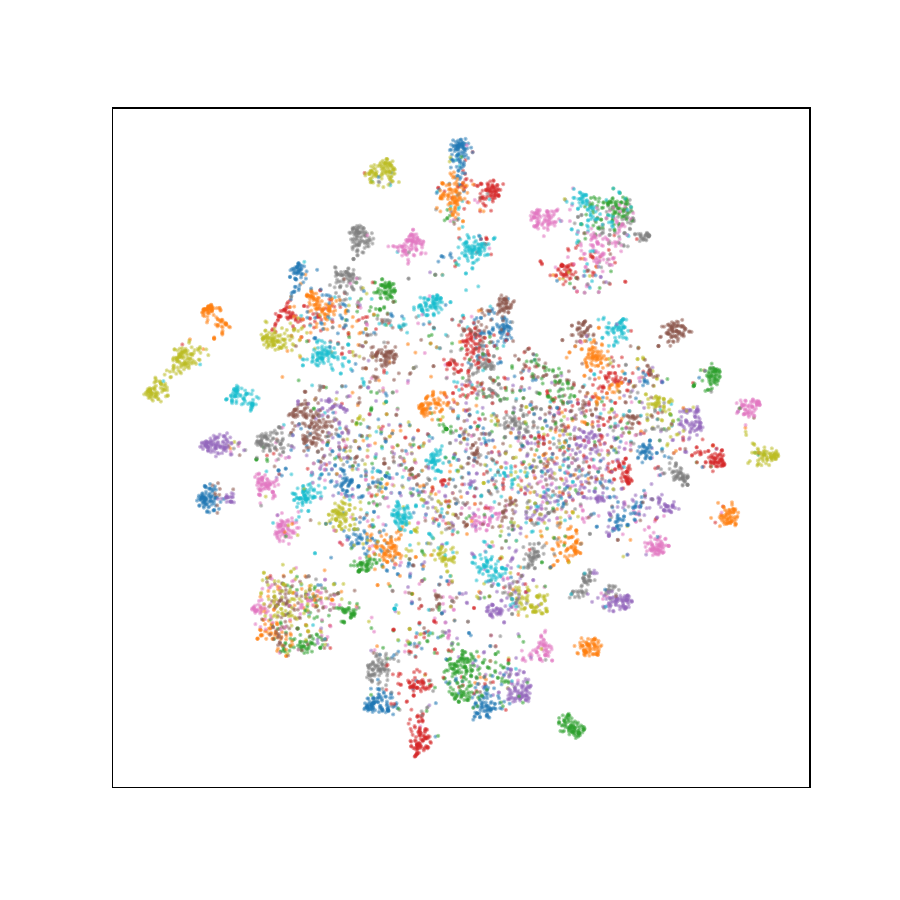}
		}
   \subfloat[DIST]{
		\includegraphics[scale=0.21]{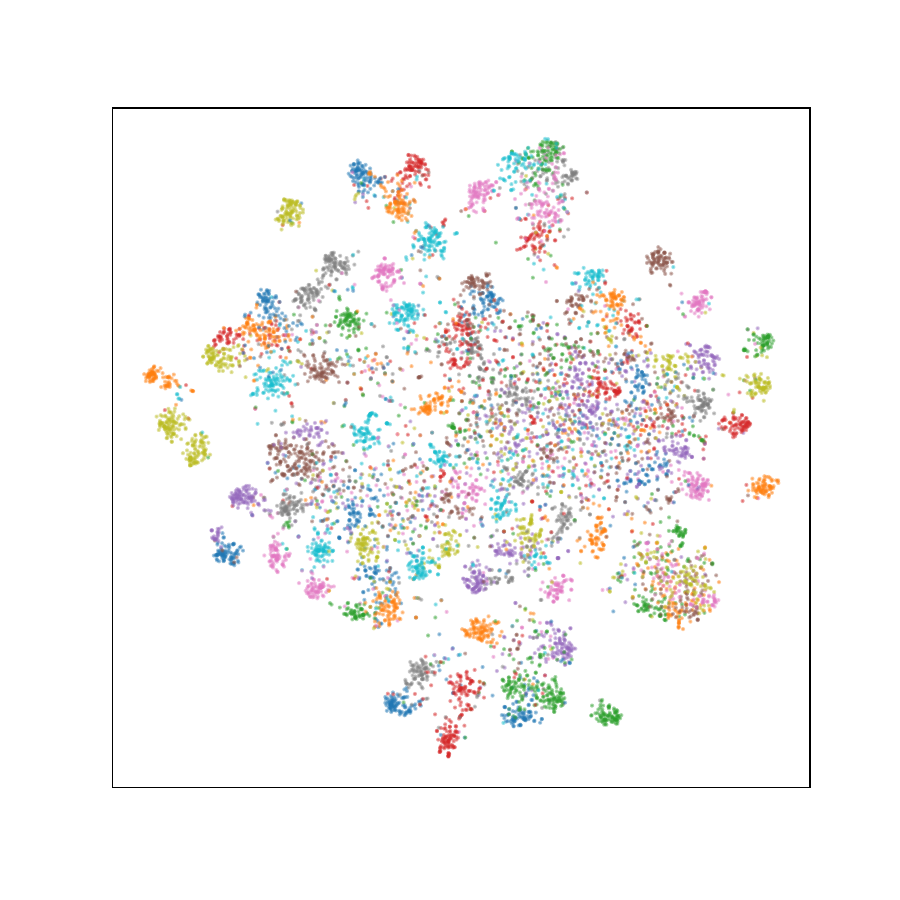}
		}
  \subfloat[\name]{
		\includegraphics[scale=0.21]{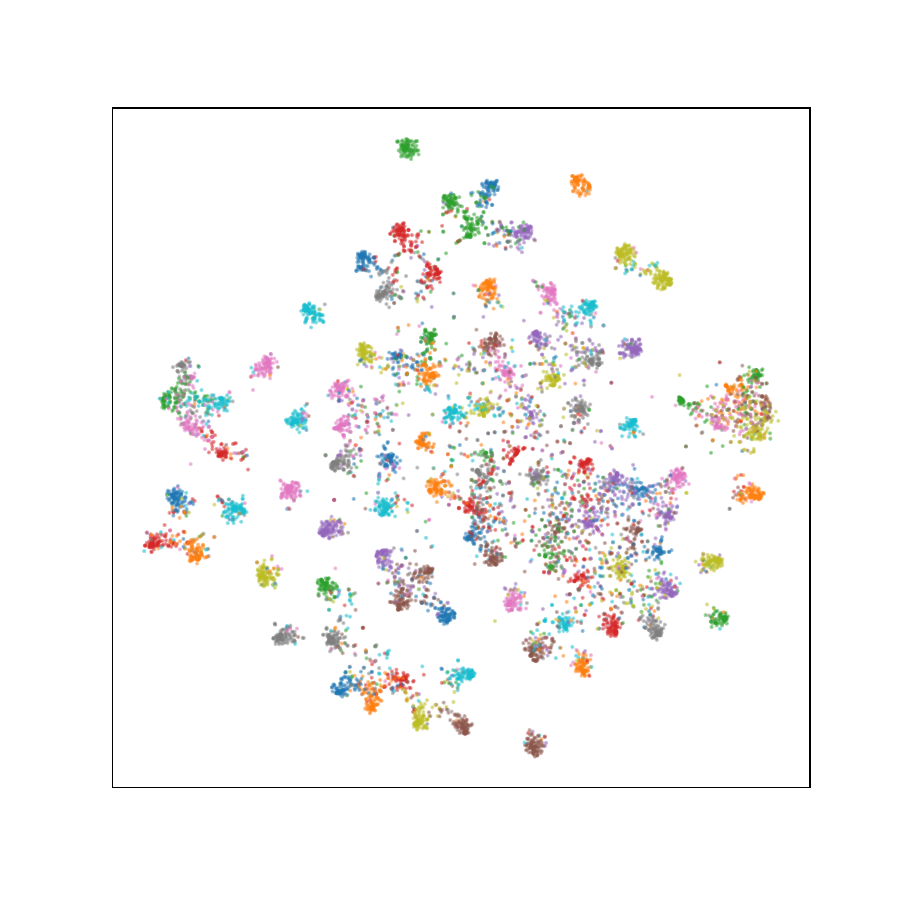}
		}
	\caption{\small t-SNE of features learned by several KD methods. We use ResNet-32$\times$4/ResNet-8$\times$4 as the teacher/student pair.}
	\label{fig: visualization on kd} 
\end{figure*}

\begin{figure}[t]
\centering
\includegraphics[width=0.88\linewidth]{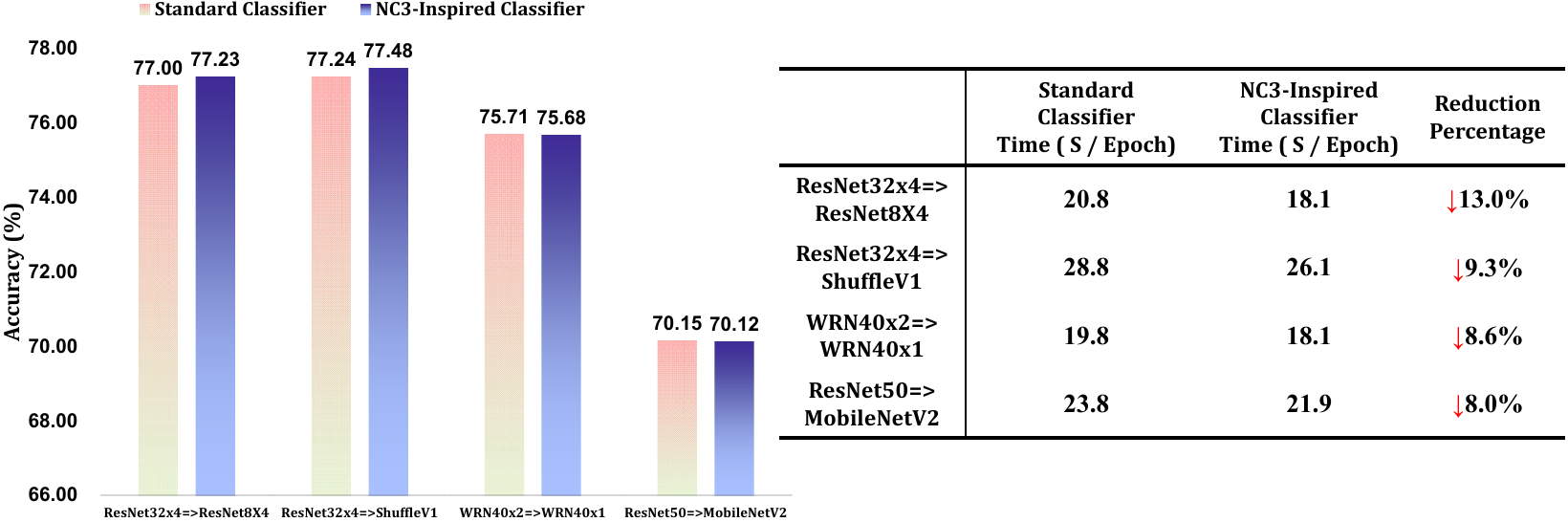} 
\caption{\small Distillation results with standard and $\mathcal{NC}_\mathbf{3}$-inspired classifiers on CIFAR-100, with training time per epoch shown in the right table.}
\label{fig:time}
\end{figure}

\section{Experiments}

\subsection{Backbones}\label{sec: basic setting}
 We compare our approach with two main kinds of KD baselines (\ie, logit-based and feature-based distillation): 
\begin{itemize}
\setlength{\itemsep}{0pt}
\setlength{\parsep}{0pt}
\setlength{\parskip}{0pt}
	\item \textbf{Logit-based} methods include the vanilla KD~\cite{KD},  
	DKD~\cite{DKD}, DIST~\cite{DIST} and MLKD~\cite{MLKD}.
	\item \textbf{Feature-based}  methods include  FitNet~\cite{Fitnets}, RKD~\cite{RKD}, PKT~\cite{PKT},   CRD~\cite{CRD}, 
    ReviewKD~\cite{ReviewKD}, FGFI~\cite{wang2019distilling}, NORM~\cite{Norm},  SimKD~\cite{SimKD}, TopKD~\cite{topkd} and TTM~\cite{TTM}.
\end{itemize}
The detailed implementation of experiments is in the Appendix. 




\subsection{Main Results}
\paragraph{CIFAR-100.}
To validate the effectiveness of our approach, we compared NCKD against a range of state-of-the-art distillation methods. Our experiments included both similar-architecture and cross-architecture distillation to demonstrate the universality of our method. As shown in \Cref{tab:cifar100-sota}, \name outperformed all existing baselines, achieving an average accuracy of 75.10\%. Additionally, when we integrated our NC-inspired losses as a plug-in module into two mainstream methods, CRD and SimKD, we observed a significant improvement in distillation performance. These results confirm the effectiveness of our approach in enhancing distillation generalization and highlight its versatility as a plug-and-play module suitable for various distillation frameworks and real-world applications.
\paragraph{ImageNet-1k.} 
To validate the effectiveness of our method on large-scale vision tasks, we conducted experiments on the ImageNet-1k dataset, using both similar-architecture (ResNet34/ResNet18) and cross-architecture (ResNet50/MobileNet) network pairs. As presented in \Cref{Tbl:ImageNet}, our method consistently outperforms the baselines, aligning with our findings on CIFAR-100. Remarkably, our approach even surpasses the advanced KD search method, DisWOT, by a substantial margin for the respective student-teacher pairs. These results highlight the effectiveness of our method in large-scale learning.


\paragraph{MS-COCO.}
We verify the efficacy of the proposed NC-inspired loss in knowledge distillation tasks for object detection on the COCO dataset, as shown in Table~\ref{tab:mscoco_results}. All methods are evaluated under uniform training conditions to ensure comparability. Specifically, \name yields a significant improvement in performance, demonstrating their effectiveness and efficiency in knowledge distillation for dense prediction tasks.

\subsection{Extensions}
\subsubsection{Visualization}
We employ t-SNE to evaluate the efficacy of our distillation method in enhancing the feature representation, as shown in \Cref{fig: visualization on kd}. KD, CRD, and DIST serve as our primary baselines. While the baseline models exhibit considerable class overlap, indicating poor feature separation, our method produces distinct clusters, demonstrating improved discriminative power. These results empirically validate the effectiveness of our approach and highlight its potential to enhance model generalization.

\subsubsection{Ablation Study} 

\paragraph{Distillation from Bigger Models.}
In principle, effective knowledge distillation should lead to \textsc{great teachers producing outstanding students}, meaning that a superior teacher should guide the student to better distillation. However, in practice, such ideal case is not always achieved. We do evaluation using ResNet and Swin models of varying scales, as shown in Table \ref{tab: different teachers}. One can observe that existing methods do not consistently guarantee steady improvements in student performance as the teacher model's size increases. In contrast, our approach effectively addresses this issue, likely because better models establish a refined NC structure, which facilitates the student's consistent enhancement.

\paragraph{Does $\mathcal{NC}$ impact KD?}
\textbf{Yes!} We evaluate the contribution of each $\mathcal{NC}$ property to the distillation process through ablation study, as shown in Table \ref{tbl:ablation}. The results show that removing any NC property would reduce the student prediction accuracy, with $\mathcal{NC}_\mathbf{2}$ having the most significant impact. This underscores the critical role of each module in our framework, especially the importance of preserving the teacher's ETF structure for effective knowledge transfer. Additionally, when combined with standard KD, our method further improves the distillation performance. 

\paragraph{Does $\mathcal{NC}_\mathbf{3}$-classifier trade performance for efficiency?}
\textbf{No!} We conduct ablation study on the $\mathcal{NC}_\mathbf{3}$ classifier, with results presented in \Cref{fig:time}. Notably, the $\mathcal{NC}_\mathbf{3}$ classifier either outperforms or matches the standard classifier's results. Additionally, as shown in the right table, 
the training time is significantly reduced, 
suggesting that this design effectively balances performance and efficiency.

\subsubsection{Case Study} 

While we are the first to explicitly integrate NC into the KD framework, we recognize that some existing methods have implicitly leveraged $\mathcal{NC}$ to enhance distillation, albeit without explicit acknowledgment. Here, we investigate the role of $\mathcal{NC}$ in the effective distillation results of two representative methods, CRD and SimKD.
\paragraph{Case 1: } CRD uses contrastive learning at the instance level to align teacher and student features, implicitly encouraging feature convergence toward class centroids~\cite{khosla2020supervised}. This is reflected in the significant reduction of $\mathcal{NC}_\mathbf{1}$ in CRD compared to KD (see Table \ref{tab:case study}), indicating its implicit use of $\mathcal{NC}_\mathbf{1}$. Our approach, however, better preserves the $\mathcal{NC}_\mathbf{1}$ property, resulting in improved  performance.
\paragraph{Case 2: } SimKD replaces the student’s classifier with the teacher’s, focusing solely on the feature matching. We hypothesize that this implicitly leverages the teacher's $\mathcal{NC}_\mathbf{3}$ property — where the reused classifier weights $\Vec{w}$ preserve the teacher's normalized centroids $\Tilde{\Vec{h}}^T$. Our calculations, shown in Table \ref{tab:case study}, indicate that SimKD achieves $\mathcal{NC}_\mathbf{3}$ values closer to 1 compared to standard KD. This suggests that SimKD gets benefit from this alignment, resulting in improved feature semantics and, consequently, better distillation outcomes.

\section{Conclusion}
In this work, we introduced a novel approach to knowledge distillation by incorporating the
structure of Neural Collapse (NC) 
into the distillation process. Our method, Neural Collapse-inspired Knowledge Distillation (NCKD), enables student models to learn not only from the teacher’s logits or features but also to emulate the geometrically elegant NC structure present in the teacher’s final-layer representations. This strategy effectively bridges the knowledge gap between teacher and student models, resulting in superior student performance. Comprehensive experiments across diverse tasks and network architectures consistently demonstrated that our method outperforms state-of-the-art distillation techniques, affirming its efficacy in enhancing both accuracy and generalization. These findings highlight the robustness and adaptability of our  NCKD, marking it a significant advancement in the field of knowledge distillation.

While our study primarily focused on distillation with a pre-trained teacher model, an unresolved area in the field is mutual distillation, where the student model also transfers knowledge back to the teacher during the distillation process. In future work, we will investigate whether NC can similarly benefit mutual distillation. Additionally, we aim to design NC-based criteria for selecting the most appropriate teacher model for a given student within the distillation framework.

\newpage

\bibliography{aaai25}
\clearpage

\appendix
\appendix

\newpage

\begin{table*}[t]
\centering
\begin{tabular}{l|c|cccccc|c}
\toprule
 & Student & KD & AT & FitNet& CRD & DIST & \name  & Teacher\\
\midrule
CIFAR100$\rightarrow$STL-10        & 71.33 & 73.01 & 73.67 & 73.12 &74.68 & 75.12  & {\bf 76.22} & 70.60 \\
CIFAR100$\rightarrow$TinyImageNet & 35.10 & 35.39 & 35.42 & 35.55 & 37.00& 37.13  & {\bf 38.58} & 34.20 \\
\bottomrule
\end{tabular}
\vspace{5pt}
\caption{
\small{We conduct the experiment of feature transfer by using the representation learned from CIFAR-100 to STL-10 and TinyImageNet datasets. We freeze the network and train a linear classifier on top of the last feature layer to perform a 10-way (STL-10) or 200-way (TinyImageNet) classification. We use the combination of teacher network ResNet-32$\times$4 and student network ResNet-8$\times$4. 
}
}
\label{tbl:transfer}
\end{table*}

\begin{table*}[t]
\setlength{\tabcolsep}{2.5pt}
\begin{center}
\begin{tabular}{l|ccccccc|cc}
    \toprule 
    Model & \small{Baseline
    }& \small{DDGSD} & \small{BYOT} & {\small{CS-KD}} & \small{SLA+SD} & \small{FRSKD} & \small{BAKE} & \small{\name} & {\color{teal} $\Delta$}\\
    \midrule 
    ResNet-50 & 76.80 & 77.10 & 77.40 & 77.61 & 77.20 & 76.68 & 78.00 & {\bf78.92} & {\color{teal} +2.12}\\
    ResNet-101 & 78.60 & 78.81 & 78.66 & 78.99 & 78.91 & 79.22 & 79.31 & {\bf79.98} & {\color{teal}+1.38} \\
    \midrule 
    ResNeSt-50 & 78.40 & 78.66 & 78.60 & 78.71 & 78.98 & 78.91 & 79.31 & {\bf80.46} & {\color{teal}+2.06} \\
    \midrule 
    ResNeXt-101 (32$\times 4d$) & 78.71 & 78.99 & 78.00 & 78.24 & 78.68 & 79.11 & 79.21 & {\bf80.23} & {\color{teal} +1.52} \\
    \bottomrule
\end{tabular}
\vspace{5pt}
\caption{Comparison of self-distillation methods on ImageNet using models of ResNet, ResNeSt and ResNeXt. The last column are the performance improvement compared to vanilla classification. {\color{teal} $\Delta$} denotes the improvement of our distillation to the baseline.}
\label{tbl:sd imagenet}
\end{center}
\vspace{10pt}
\end{table*}

\begin{figure*}
    \centering
    \includegraphics[width=0.98\linewidth]{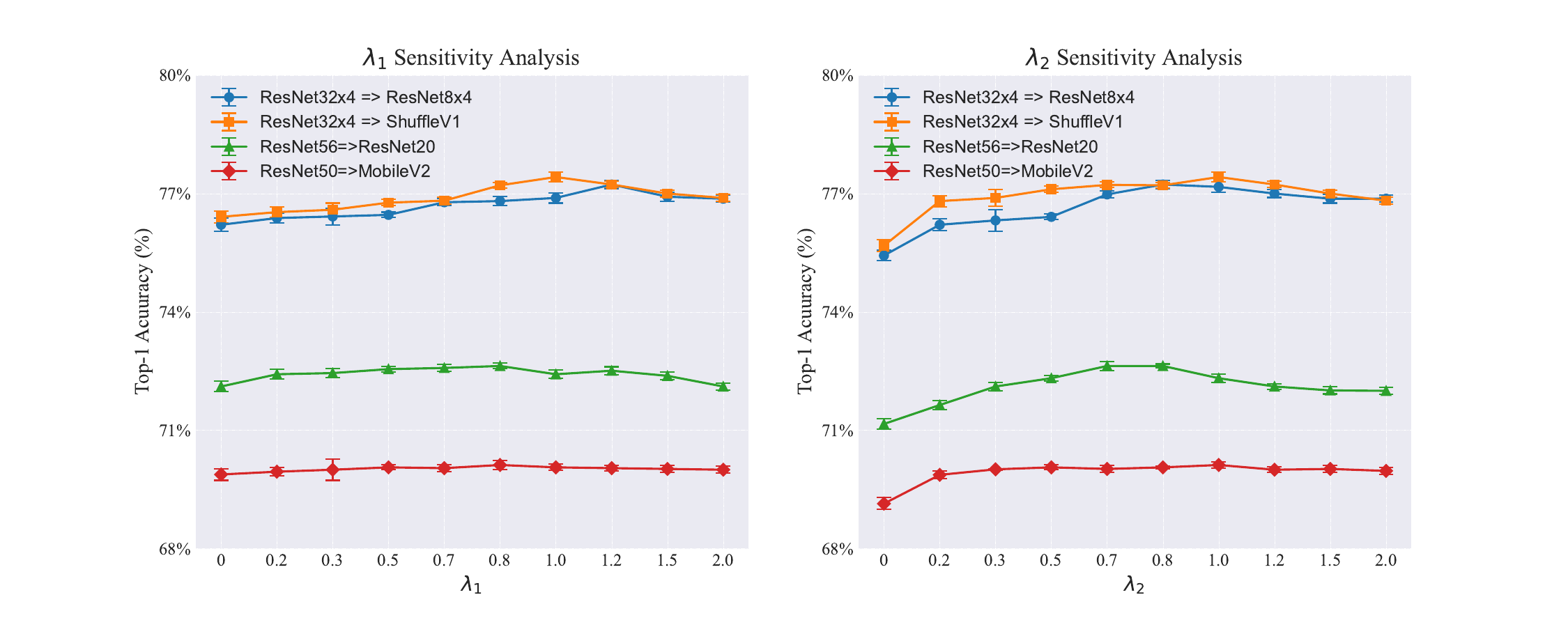}
    \caption{Sensitivity analysis on hyper-parameters $\lambda_1, \lambda_2$. All experiments were conducted on the CIFAR-100 dataset, with each experiment repeated three times. The mean and standard deviation of the results are presented in the figures.}
    \label{fig:sensitivity}   
\end{figure*}

\begin{figure}[t]
    \centering
    \includegraphics[width=0.92\linewidth]{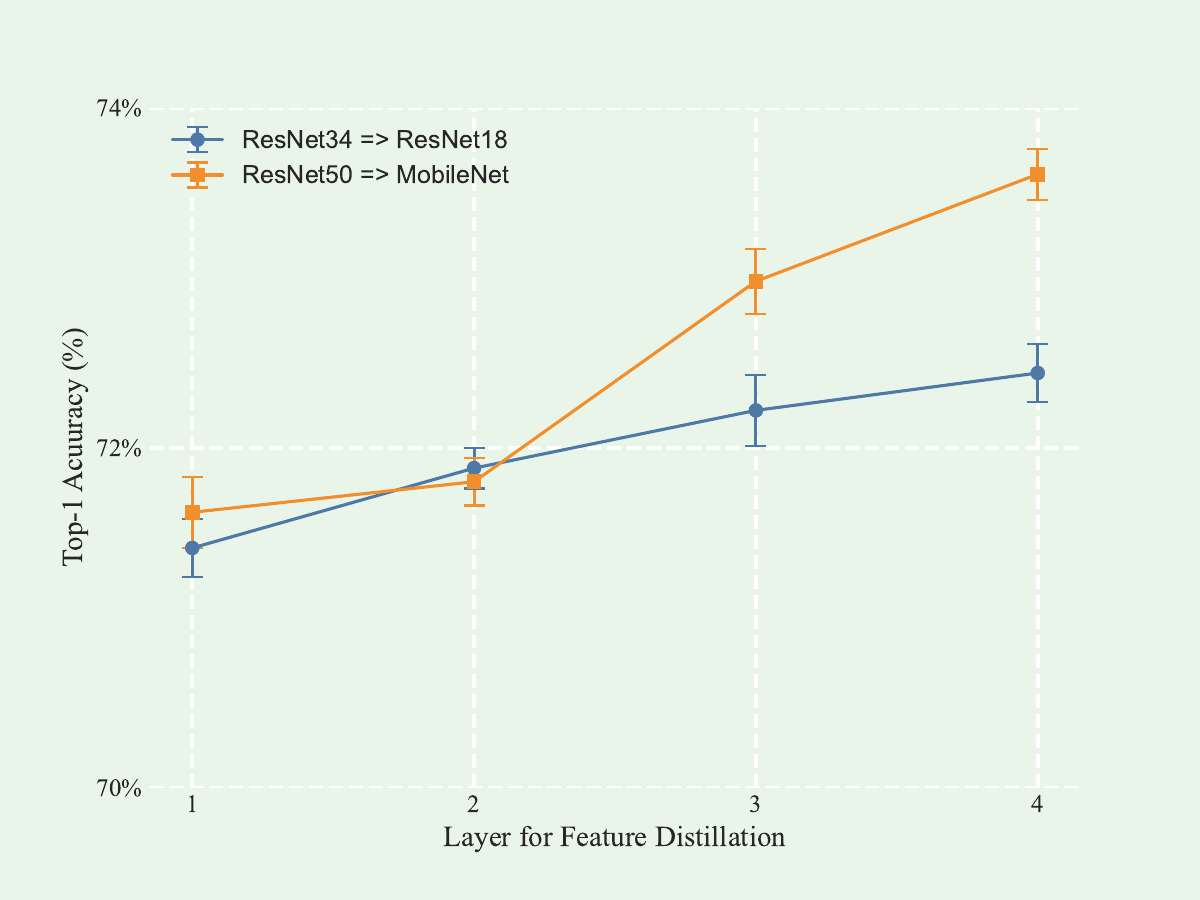}
    \caption{Ablation Study on the layer for \name. All experiments were conducted on the ImageNet-1k.}
    \label{fig:layer}   
\end{figure}

\section*{A. Detailed Explanation of $\mathcal{NC}_\mathbf{2}$ Distillation}
\begin{proof}
    Intuitively, for the $i$-th normalized student prototype $\Tilde{\Vec{h}}_i^S$ to replace the corresponding teacher centroid $\Tilde{\Vec{h}}_i^T$ and form a simplex Equiangular Tight Frame (ETF) structure with the other centroids of the teacher, it must satisfy the following conditions:
    \begin{equation}
        \begin{aligned}
            &\Tilde{\Vec{h}}_i^S \cdot \Tilde{\Vec{h}}_i^T = 1, \\
            & \Tilde{\Vec{h}}_i^S \cdot \Tilde{\Vec{h}}_j^T = -\frac{1}{K-1} \quad \text{for}  \quad i \neq j.
        \end{aligned}
    \end{equation}
Thus, the inner product between $\Tilde{\Vec{h}}_i^S$ and teacher's prototypes $\Tilde{\Vec{H}}^T = [\Tilde{\Vec{h}}_1^T, \cdots, \Tilde{\Vec{h}}_i^T, \cdots, \Tilde{\Vec{h}}_j^T, \cdots, \Tilde{\Vec{h}}_K^T]$ has the form  $\Tilde{\Vec{h}}_i^S \cdot \Tilde{\Vec{H}}^T = [-\frac{1}{K-1}, \cdots, 1, \cdots, -\frac{1}{K-1}, \cdots, -\frac{1}{K-1}]$. To align all $\Tilde{\Vec{h}}_i^S \cdot (\Tilde{\Vec{h}}^T)^\top$, we have:
\begin{equation}
    \Tilde{\Vec{H}}^S ({\Tilde{\Vec{H}}^T})^\top = \begin{pmatrix}
  1& -\frac{1}{K-1} & \cdots & \cdots & -\frac{1}{K-1} \\
  -\frac{1}{K-1} & 1 & \cdots & \cdots & \cdots \\
 \cdots & \cdots & 1 & \cdots &\cdots \\
  \cdots& \cdots & \cdots & \cdots & \cdots \\
  -\frac{1}{K-1}& \cdots & \cdots & \cdots &1 
\end{pmatrix} .
\label{eq:inner product}
\end{equation}
Noting that the $\Vec{I}_K - \frac{1}{K}{\bf 1}_K {{\bf 1}_K}^\top$ term from \cref{eq:NC2} has the form of:
\begin{equation}
    \Vec{I}_K - \frac{1}{K}{\bf 1}_K {{\bf 1}_K}^\top = \begin{pmatrix}
  \frac{K-1}{K}& -\frac{1}{K} & \cdots & \cdots & -\frac{1}{K} \\
  -\frac{1}{K} & \frac{K-1}{K} & \cdots & \cdots & \cdots \\
 \cdots & \cdots & \frac{K-1}{K} & \cdots &\cdots \\
  \cdots& \cdots & \cdots & \cdots & \cdots \\
  -\frac{1}{K}& \cdots & \cdots & \cdots &\frac{K-1}{K}
\end{pmatrix}. 
\label{eq:expansion}
\end{equation}
Using $\frac{K}{K-1}$ to multiply \cref{eq:expansion}, we have:
\begin{small}
    \begin{equation}
    \frac{K}{K-1}(\Vec{I}_K - \frac{1}{K}{\bf 1}_K {{\bf 1}_K}^\top) = \begin{pmatrix}
  1& -\frac{1}{K-1} & \cdots & \cdots & -\frac{1}{K-1} \\
  -\frac{1}{K-1} & 1 & \cdots & \cdots & \cdots \\
 \cdots & \cdots & 1 & \cdots &\cdots \\
  \cdots& \cdots & \cdots & \cdots & \cdots \\
  -\frac{1}{K-1}& \cdots & \cdots & \cdots &1
\end{pmatrix}.
\label{eq:comprise}
\end{equation}
\end{small}
By integrating \cref{eq:comprise,eq:inner product}, it can be inferred that optimizing the loss function \cref{eq:NC2} leads to a geometric alignment between the student and teacher models in terms of the NC structure, thereby improving the distillation process. This completes the proof.
\end{proof}

\section*{B. Experimental Settings}
\paragraph{CIFAR-100}~\cite{CIFAR} comprises $32\times32$
pixel color images representing objects from 100 distinct categories. For the fair comparison,we follow the standard practice in ~\cite{CRD}. We train the student network using a mini-batch size of 128 over 240 epochs, employing SGD with a weight decay of 5e-4 and momentum of 0.9. The initial learning rate is set to 0.1 for ResNet~\cite{resnet} and WRN~\cite{zagoruyko2016wide} backbones, and 0.01 for lightweight MobileNet~\cite{mobilenetv2} and ShuffleNet~\cite{zhang2018shufflenet} backbones, decaying it with a factor of 10 at 150-th, 180-th, and 210-th.  The temperature is empirically set to 4 for KD-related~\cite{KD} methods. All hyper-parameters $\lambda_1$ and $\lambda_2$ are chosen using grid search from the range of $[0,2]$, we set $\tau$ as $0.1$ following the practice of CRD~\cite{CRD}. 

\paragraph{ImageNet}~\cite{ImageNet} comprises 1.28 million images for training and 50,000 images for validation, spanning 1,000 diverse object categories. For our evaluation, we follow the standard augmentation~\cite{CRD} using pre-processed images (resized to 256x256 and cropped to 224x224, normalized with ImageNet mean/std).  We employ SGD with a mini-batch size of 512 for a total of 120 epochs (with a linear warmup for the first 5 epochs). The initial learning rate is set to 0.2 and is reduced by a factor of 10 every 30 epochs. Besides, the weight decay and momentum are set to 1e-4 and 0.9, respectively. We also expand the investigation to include the impact of distillation from large pre-trained models such as BiT~\cite{BiT} and Swin~\cite{Swin}, beyond the basic network configurations. We directly use the optimal hyper-parameters selected from CIFAR-100 as the default set. All ImageNet experiments are performed on 4 RTX 3090 GPUs, with the total epochs set at 120, focusing on maximizing top-1 accuracy in the validation set. The pre-trained weights for teachers come from PyTorch\footnote{https://pytorch.org/vision/stable/models.html}  for fair comparisons.

\paragraph{COCO 2017}~\cite{coco} comprises 118k training images and 5k validation images across 80 categories. We utilize Faster R-CNN~\cite{ren2015faster} with FPN~\cite{lin2017feature} as the feature extractor, wherein both teacher and student models adopt ResNet~\cite{resnet,resnetv1} as the backbone. In addition, MobileNet-V2 is used to evaluate cross-architecture distillation. All student models are trained with 1x scheduler, following Detectron2~\footnote{https://github.com/facebookresearch/detectron2}.

\section*{C. More Experiments}
\subsection*{C.1. Feature Transfer}
We also wonder to know whether the network distilled using NCKD exhibits feature transfer capabilities. Therefore, we continue to conduct several experiments to examine the feature transferability of \name.
As shown in \Cref{tbl:transfer}, we train linear fully-connected (FC) layers as the classifier with the feature extractor frozen for STL-10~\cite{coates2011analysis} and Tiny-ImageNet~\cite{Le2015TinyIV} datasets. We use an SGD optimizer with 0.9 momentum and no weight decay strategy in classifier training. We set the batch size to 128, and the number of total epochs to 40. Our initial learning rate is set to 0.1, then divided by 10 for every 10 epochs. From \cref{tbl:transfer}, we observe that our method beats all existing techniques, manifesting its feature transferability. 

\subsection*{C.2 Self-Distillation}
To further validate the effectiveness of \name in teacher-free distillation scenarios, we adopt the teacher-free distillation framework introduced in CS-KD \cite{CS-KD} and modify its original loss function with our newly proposed loss function, as defined in \cref{eq:total}. Within this framework, the network is encouraged to utilize features to form a simplex ETF structure, achieving self-alignment with its own Neural Collapse (NC) structure. We assess the performance of \name against various prominent teacher-free distillation methods (including DDGSD~\cite{xu2019data}, BYOT~\cite{zhang2019your},CS-KD, SLA~\cite{lee2020self},FRSKD~\cite{ji2021refine}, BAKE~\cite{ge2021self}) on the ImageNet dataset. As illustrated in Table \ref{tbl:sd imagenet}, our approach outperforms other self-knowledge distillation baselines on ImageNet, not only with commonly used architectures such as ResNet (e.g., ResNet-50) but also when applied to ResNeSt \cite{zhang2022resnest} and ResNeXt \cite{xie2017aggregated} networks.
This suggests that our approach remains effective within the teacher-free paradigm.

\section*{D. Additional Ablation Studies}

\subsection*{D.1 Sensitivity Analysis}
We evaluate the impact of the hyper-parameters $\lambda_1$ and $\lambda_2$ of \cref{eq:total} on the results, which are presented in \Cref{fig:sensitivity}. It is observed that both hyper-parameters exhibited a trend of initially decreasing and then increasing performance. Moreover, both graphs demonstrate that when the parameters exceed 0, the prediction accuracy begins to improve. This highlights the indispensability of our two distillation loss components. Additionally, it is noteworthy that the sensitivity curve of $\lambda_2$ is steeper, indicating the significant role of learning the teacher's refined NC structure in the distillation process.

\subsection*{D.2 The Impact of Layer Choice for Distillation}
Given that neural networks typically exhibit a pronounced NC structure in the features of the final layer, while earlier layers do not exhibit this structure as clearly, we investigate the impact of selecting different layers' features for \name. As shown in \Cref{fig:layer}, we observe a monotonic improvement in distillation performance as deeper layers are selected. This aligns with our expectations, as deeper layers are better able to capture the NC structure, thereby utilizing NC information to enhance the effectiveness of the distillation process.

\end{document}